# Development of a Canada-Wide Morphology Map for the ITU-R P. 1411 Propagation Model

Jennifer P. T. Nguyen

Communications Research Centre Canada
3701 Carling Avenue, Ottawa, ON Canada

*Abstract*— **This paper outlines the development of a Canada-wide morphology map classifying regions into residential, urban low-rise, and urban high-rise environments, following the ITU-R P.1411-12 propagation model guidelines. To address the qualitative nature of the environment-type descriptors found in the Recommendation, a machine learning approach is employed to automate the classification process. Extensive experimentation optimized classification accuracy, resulting in a Canada-wide morphology map that ensures more accurate path loss estimations for outdoor short-range propagation at frequencies ranging from 300 MHz to 100 GHz.**

## I. INTRODUCTION

The ITU-R P.1411-12 propagation model applies to outdoor short-range propagation for frequencies ranging from 300 MHz to 100 GHz, applicable to both line-of-sight and non-line-of-sight environments [1]. It provides a recommended method of calculating path loss across various scenarios. To identify the scenario effectively, one of the input parameters is the environment type.

Currently, there is no Canada-wide morphology map that classifies regions according to the ITU-R P.1411 environment types. Existing city-specific maps, used for purposes like city planning and land use statistics, vary in methodology and lack consistency, making them unsuitable for nationwide application. Therefore, this work develops a comprehensive Canada-wide morphology map categorizing regions into residential (RES), urban low-rise (ULR), and urban high-rise (UHR) environments, as defined by the ITU-R P.1411 propagation model recommendations.

The ITU-R P.1411 environment categories are qualitatively defined, necessitating a consistent classification method to establish a reliable "ground truth." Rather than manually labelling each area, a subset of city regions is used as training data for a machine learning (ML) algorithm, automating the classification process. This approach enables the generation of a morphology map for the 92 Canadian cities with populations over 30,000, using standardized and reproducible criteria to ensure alignment with the ITU guidelines.

## II. RECOMMENDATION ITU-R P. 1411 AND ITS APPLICABILITY

To select the appropriate P.1411 scenario, identifying the environment type of the region in which each base station pair (transmitter and receiver) is installed is essential. For this reason, a morphology map is crucial for selecting the appropriate environmental scenario. When two base stations are located in different environments, the scenario resulting in the lowest path loss should be used to compute interference.

In the site-general scenarios of the P. 1411 propagation model [1], the environment types considered are urban very high-rise, urban high-rise, urban low-rise/suburban, and residential. In this work, urban very high-rise and urban high-rise environments are combined into a single category. The propagation impairments and typical mobile velocity for each environment category are described in Tables I and II, respectively.

TABLE I. PHYSICAL OPERATING ENVIRONMENTS AND PROPAGATION IMPAIRMENTS [1]

| Physical operating environments and propagation impairments | |
|---|---|
| *Environment* | *Description and propagation impairments of concern* |
| Urban high-rise | • Urban canyon, characterized by streets lined with tall buildings of several floors each<br>• Building height makes significant contributions from propagation over roof-tops unlikely<br>• Rows of tall buildings provide the possibility of long path delays<br>• Large numbers of moving vehicles in the area act as reflectors adding Doppler shift to the reflected waves |
| Urban low-rise/ Suburban | • Building heights are generally less than three stories making diffraction over roof-top likely<br>• Reflections and shadowing from moving vehicles can sometimes occur<br>• Primary effects are long delays and small Doppler shifts |
| Residential | • Single and double storey dwellings<br>• Roads are generally two lanes wide with cars parked along sides<br>• Heavy to light foliage possible<br>• Motor traffic usually light |

TABLE II. PHYSICAL OPERATING ENVIRONMENTS AND TYPICAL MOBILE VELOCITY [1]

| Physical operating environments and typical mobile velocity | |
|---|---|
| *Environment* | *Velocity for vehicular users* |
| Urban high-rise | Typical downtown speeds around 50 km/h (14 m/s) |
| Urban low-rise/Suburban | Around 50 km/h (14 m/s) Expressways up to 100 km/h (28 m/s) |
| Residential | Around 40 km/h (11 m/s) |

## III. RELATED WORK

### A. Existing Datasets

Several open-source datasets classify regions within Canada for various purposes. Many cities maintain their own land-use maps for urban planning, but these maps lack a standardized format. Therefore, they cannot be directly mapped to the environment-type descriptors used for the ITU P.1411 propagation model. One example of such a dataset is the *'Quartier de référence en habitation'* [2] geographic dataset that divides Montreal into categories needed for a housing analysis.

Another notable dataset is the Canadian Suburbs Atlas [3], which uses 2021 census data to categorize Canadian suburbs into three types: active core, transit suburbs, and automobile suburbs. This classification provides valuable insights for urban planning, particularly regarding connectivity and infrastructure within suburban areas. While the Canadian Suburbs Atlas broadly defines suburban areas to include residential zones, this classification differs from the ITU P.1411 terminology.

### B. Literature Review of Similar Works

Several studies have classified building vector data into various categories, often using methodologies where researchers define the environment types based on their specific objectives. This results in varying numbers and definitions of categories. The initial categorization is often executed through visual classification, which serves as the training data. For example, [4] uses a classification approach that is based only on the characterisation of building geometries with morphological measures derived simply from human perception and pattern recognition. Many approaches utilize free and open-source geospatial data, supplemented with additional datasets containing points of interest.

These methodologies typically involve defining building features and extracting relevant metrics to evaluate their importance. Data is often sampled on a geospatial dataset at a fixed interval with smoothing around a buffer radius to ensure consistent classification across the dataset.

## IV. METHODOLOGY

### A. Overview of Pipeline

The methodology proposed here for generating the morphology map is as follows:
1. Prepare the geospatial data, which includes both building and road network data.
2. Select training area polygons through visual inspection to align with the ITU recommendations and later compared with existing datasets, only including unambiguous areas.
3. Execute per-location calculations, or point-based neighbourhood analysis, for both training and testing areas.
4. Apply a ML classification algorithm using the training data to identify areas with similar characteristics (based on feature importance) in other cities and generate the morphology map.
5. Analyze the accuracy of the predictions and compare the newly generated morphology maps with existing similar maps.

### B. Geospatial Data Preparation

The process begins with preparing the geospatial data, encompassing building geodata and road network data. This geospatial data contains building vectors with height information for each of the 92 Canadian cities. Convex hulls are created around the geodata to serve as boundaries for subsequent processing. Road network data from Open Street Map (OSM) is retrieved to extract specified road classifications (motorway, trunk, primary, secondary, tertiary, unclassified, and residential). The extracted attributes include the number of lanes, maximum speed for each road, and road names for reference. These results are exported as shapefiles, retaining only the features specified in the P.1411 recommendations.

### C. Defining Ground Truth

Initially, regions are manually identified as residential, urban low-rise, and urban high-rise through satellite imagery inspection. These areas are then validated using Google Maps Street View to ensure that they meet the descriptor criteria defined for each environment type in the ITU-R recommendations. Only regions that unambiguously match these descriptions are selected and labelled to ensure alignment with ITU guidelines. Figure 1 illustrates the areas selected for training in Montreal, Canada, as the coloured polygons.

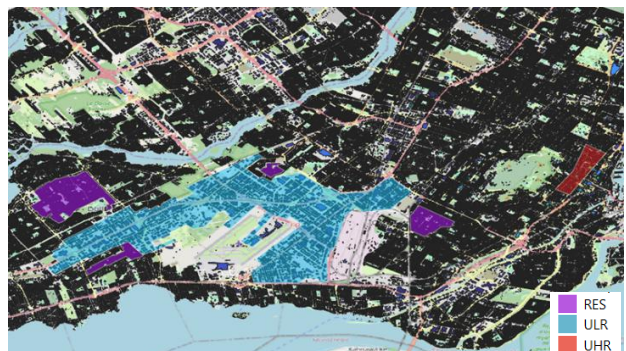

Figure 1. Selected areas for training in Montreal

### D. Per-Location Calculations

Grid points are generated within the training polygons shown in Figure 1 at 30-metre intervals. For each point in the training polygons, features from Table III are extracted within a 300-metre buffer radius to quantify the environment-type descriptions. Each training point is labelled according to the environment type of its corresponding polygon. This process establishes relevant statistics and observes trends to provide essential insights for data standardization and defining the ground truth. Additionally, obtaining averages within a buffer radius as opposed to directly sampling the geospatial data minimizes the effect of outlier buildings on the machine learning model.

TABLE III. STATISTICS OF MONTREAL TRAINING DATA

| Statistics of Montreal Training Data | | | |
|---|---|---|---|
| *Features* | *Residential* | *Urban Low-Rise* | *Urban High-Rise* |
| Average building heights (m) | 6.45 | 6.94 | 20.25 |
| Median building heights (m) | 6.45 | 6.81 | 16.75 |
| Standard deviation of building heights (m) | 1.81 | 2.95 | 15.76 |
| Maximum building height (m) | 15.17 | 12.65 | 65.62 |
| Minimum building height (m) | 1.49 | 2.38 | 0.76 |
| Average building area (m$^2$) | 192.91 | 5485.19 | 2327.38 |
| Maximum building area (m$^2$) | 2534.24 | 21350.47 | 17639.33 |
| Minimum building area (m$^2$) | 11.55 | 550.27 | 8.74 |
| Total building count | 205.67 | 19.22 | 49.4 |
| Footprint density | 0.13 | 0.23 | 0.38 |
| Average number of lanes | 1.99 | 2.26 | 1.84 |
| Average speed limit (km/h) | 39.96 | 48.89 | 34.18 |

### E. Machine Learning Algorithm

The choice of classification, as opposed to clustering, aligns with ITU's predefined environment types and ensures adherence to P.1411 guidelines. Classification involves assigning each point to a predefined category based on the specific features identified through ML. This differs from clustering, which identifies similarities between points and groups them into categories without considering predefined labels. Clustering could result in the creation of new, overly specific environment type categories, potentially tailored to a single city if the training data primarily derives from there; this would disregard the ITU P.1411 recommendations in favour of aligning with points from the training dataset. As supported by the literature, classification is the most suitable method for this work [4].

The Random Forest Classifier ML algorithm is used due to its robustness and lower likelihood of overfitting compared to simple decision trees. Random Forest constructs a set of decision trees using different random subsets of data and features. Each tree casts a vote on the prediction, and for classification problems, the final output is determined through a majority vote. This work uses the Scikit-learn Python library [5] for classification.

Figure 2 shows the most important features found based on the labelled training data. As illustrated, the total building count and the average building area within the 300-metre buffer radius are the two most important features. In other words, these features are best at differentiating the three environments from each other. This figure also shows that other features, such as the number of lanes and the minimum building height, have little importance on the ML model performance.

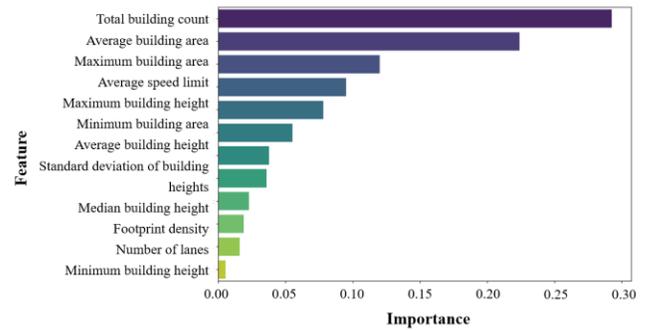

Figure 2. Feature importance

## V. PERFORMANCE EVALUATION

### A. Results

The results obtained using Random Forest demonstrate an accuracy of 0.9995 when the labelled data is split into 70% training and 30% testing. Table IV shows the confusion matrix resulting from this ML model, demonstrating near-perfect precision and recall, and Table V shows the derived precision and recall of this ML model.

TABLE IV. CONFUSION MATRIX OF MACHINE LEARNING MODEL

| Confusion Matrix (Number of Points from the Labelled Data) | | | |
|---|---|---|---|
| | *Residential* | *Urban Low-Rise* | *Urban High-Rise* |
| **Residential** | 3901 | 7 | 0 |
| **Urban Low-Rise** | 5 | 17893 | 0 |
| **Urban High-Rise** | 0 | 0 | 913 |

TABLE V. PRECISION AND RECALL OF MACHINE LEARNING MODEL

| | *Precision* | *Recall* |
|---|---|---|
| **Residential** | 0.9987 | 0.9982 |
| **Urban Low-Rise** | 0.9996 | 0.9997 |
| **Urban High-Rise** | 1.0000 | 1.0000 |

Figure 3 shows the resulting Montreal morphology map where all training points (contained within the coloured outlines) are shown to be correctly identified as their respective environment type.

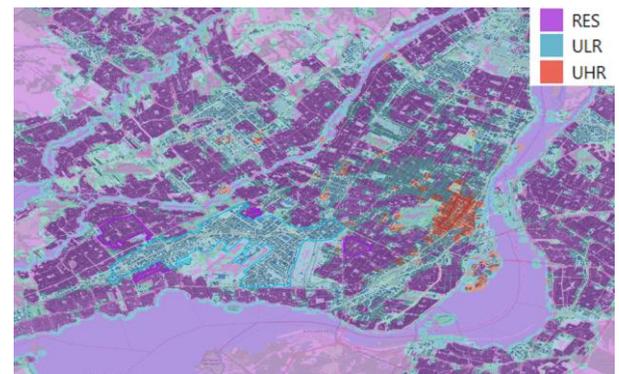

Figure 3. Morphology map for Montreal

The ML model trained on all labelled data is then applied to other Canadian cities.

### B. Analysis and Validation

By varying the per-location calculations' buffer radius, the impact on metrics such as building height and density is tested and analyzed. This experimentation informed the optimal radius choice, improving classification consistency.

Visual and statistical checks are conducted to validate the model. The model's classifications are also compared with local urban planning maps for accuracy. This comparison is both qualitative and quantitative using statistics to find similarities in the environment type assigned for each point between the maps.

The newly generated morphology maps provided clearer distinctions in urban high-rise areas, with noticeable improvements in classifying urban high-rise and low-rise zones, particularly where previous maps used more arbitrary cutoffs, such as defining residential areas as areas where building height is less 10 metres.

Decision boundary plots help to understand how the ML model predicts the environment types. Figure 4 illustrates the decision boundaries using average building height and total building count. It is just one of many decision boundaries. The top and bottom 1% of points were excluded for visualization purposes, though they are still considered in the analysis.

As seen in Figure 4, the total building count, being the most important feature, effectively classifies residential areas, while the average building height helps to distinguish urban high-rise areas as being above 10 metres. Urban low-rise areas are harder to differentiate using only building height. The background colour of this decision boundary indicates the predicted classification based on these two features alone, while the point colours reflect the final classification from the Random Forest model.

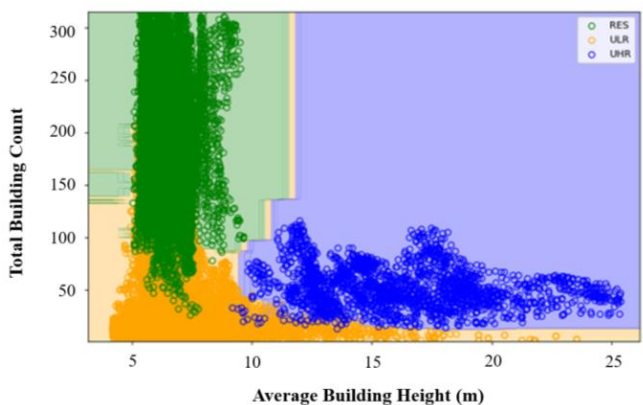

Figure 4. Decision boundaries for the average building height versus total building count in Montreal

### C. Future Work

Classification of the P. 1411 environment types can be improved in various ways as city data is updated in different geospatial datasets, ensuring alignment with recent developments. The methodology can be enhanced by incorporating more training areas, particularly in cities with distinct architectural norms for building sizes and density, even within the same environment type. For instance, residential buildings may vary significantly across different cities.

This methodology can also be adapted to incorporate additional training data over time, with validation using different urban planning datasets from various cities, ensuring continued improvements to classification accuracy.

A key next step involves comparing the path loss obtained using the environment type selected by the ML-generated morphology map with real-world path loss measurements. This would allow for cross-validation of the morphology map with measurement data, allowing for further confirmation of the morphology map performance accuracy. It is important to note that the morphology maps will still be generated by strictly following the P. 1411 recommendations and not by developing the maps around measurement data.

## VI. CONCLUSION

This paper demonstrated that the ML classification methodology effectively distinguishes ITU-recommended environment types. Through experimentation and systematic analysis, the resulting morphology maps align with P.1411 guidelines, offering clarity in areas previously constrained by ambiguous descriptors. This adaptable Canada-wide morphology map ensures more accurate path loss estimations for outdoor short-range propagation at frequencies ranging from 300 MHz to 100 GHz, as supported by the P. 1411 recommendations.

ACKNOWLEDGMENT

The author would like to express her gratitude to research colleagues, particularly Joey Wang and students Nakul Parekh and Mahad Ahmed, at the Communications Research Centre Canada for their helpful contribution to this work.

REFERENCES

[1] Report ITU-R P.1411-12: Propagation data and prediction methods for the planning of short-range outdoor radiocommunication systems and radio local area networks in the frequency range 300 MHz to 100 GHz.

[2] "Quartiers de référence en habitation – Données Québec," Donneesquebec.ca, 2017. Available: https://www.donneesquebec.ca/recherche/dataset/vmtl-quartiers

[3] D. Gordon and R. Herteg, "Canadian Suburbs Atlas," Jun. 2023. Available:https://www.canadiansuburbs.ca/wp-content/uploads/2023/10/Canadian_Suburbs_Atlas_v14min.pdf

[4] S. Steiniger, T. Lange, D. Burghardt, and R. Weibel, "An Approach for the Classification of Urban Building Structures Based on Discriminant Analysis Techniques," Transactions in GIS, vol. 12, no. 1, pp. 31–59, Feb. 2008, doi: https://doi.org/10.1111/j.1467-9671.2008.01085.x.

[5] F. Pedregosa et al., "Scikit-learn: Machine Learning in Python," Journal of Machine Learning Research, vol. 12, no. 85, pp. 2825–2830, 2025. Available: https://jmlr.csail.mit.edu/papers/v12/pedregosa11a.html